\documentclass[sigconf]{acmart}

\renewcommand\footnotetextcopyrightpermission[1]{} 

\usepackage{epsfig}
\usepackage{graphicx}
\usepackage{amsmath}
\usepackage{amssymb}

\usepackage[font=footnotesize,labelfont=bf]{caption}
\usepackage{multirow}
\usepackage{tabularx}
\usepackage{dsfont}
\usepackage{url}
\usepackage{color}
\usepackage{subfig}
\usepackage{amsthm}
\usepackage[export]{adjustbox}
\usepackage{comment}

\usepackage[utf8]{inputenc} 
\usepackage[T1]{fontenc}    

\usepackage{url}            
\usepackage{booktabs}       
\usepackage{amsfonts}       
\usepackage{nicefrac}       
\usepackage{microtype}      
\usepackage{color}
\usepackage{colortbl}

\usepackage{svg}            

\usepackage{enumitem}

\definecolor{dg}{rgb}{0,0.694,0.298}
\definecolor{purple}{rgb}{0.4,0.176,0.569}

\usepackage{pifont}

\newcommand{\qichange}[1]{\textbf{\textcolor{orange}{#1}}}

\definecolor{tabgray1}{rgb}{0.5,0.5,0.5}
\definecolor{tabgray2}{rgb}{0.85,0.85,0.85}
\definecolor{top1}{rgb}{1.0, 0.6, 0.6} 
\definecolor{top2}{rgb}{0.94, 0.9, 0.55}
\definecolor{top1-2}{rgb}{1.0, 0.66, 0.66} 
\definecolor{top1-3}{rgb}{1.0, 0.72, 0.72} 
\definecolor{top1-4}{rgb}{1.0, 0.78, 0.78} 
\definecolor{top1-5}{rgb}{1.0, 0.84, 0.84} 
\definecolor{top1-6}{rgb}{1.0, 0.90, 0.90} 
\definecolor{top1-7}{rgb}{1.0, 0.96, 0.96} 

\usepackage{xspace}
\makeatletter
\DeclareRobustCommand\onedot{\futurelet\@let@token\@onedot}
\def\@onedot{\ifx\@let@token.\else.\null\fi\xspace}
\def\eg{\emph{e.g}\onedot} 
\def\ie{\emph{i.e}\onedot} 
 
\def\etc{\emph{etc}\onedot} 
 
\def\etal{\emph{et al}\onedot}
\makeatother

\AtBeginDocument{%
	\providecommand\BibTeX{{%
			\normalfont B\kern-0.5em{\scshape i\kern-0.25em b}\kern-0.8em\TeX}}}

\copyrightyear{2020} 
\acmYear{2020} 
\setcopyright{acmcopyright}\acmConference[MM '20]{Proceedings of the 28th ACM International Conference on Multimedia}{October 12--16, 2020}{Seattle, WA, USA}
\acmBooktitle{Proceedings of the 28th ACM International Conference on Multimedia (MM '20), October 12--16, 2020, Seattle, WA, USA}
\acmPrice{15.00}
\acmDOI{10.1145/3394171.3413707}
\acmISBN{978-1-4503-7988-5/20/10}

\begin{document}
	\fancyhead{}
	
	\title{\emph{DeepRhythm}: Exposing DeepFakes with Attentional Visual Heartbeat Rhythms}
	
	\author{Hua Qi$^{1*}$, \ Qing Guo$^{2*}$, \ Felix Juefei-Xu$^{3}$, \ Xiaofei Xie$^{2}$, \ Lei Ma$^{1{\dagger}}$}
	\author{\ Wei Feng$^{4}$, \ Yang Liu$^{2}$, \ Jianjun Zhao$^{1}$}
	
	\thanks{$^{*}$ Both authors contributed equally to this research. \\
		$^{\mathrm{\dagger}}$ Lei Ma is the corresponding author (malei@ait.kyushu-u.ac.jp).}
	\affiliation{\institution{
			$^{1}$Kyushu University, Japan \ \ $^{2}$Nanyang Technological University, Singapore \\ }}
	\affiliation{\institution{$^{3}$Alibaba Group, USA \ \ $^{4}$Tianjin University, China}}
	
	\renewcommand{\shortauthors}{Hua Qi et al.}
	
	\begin{abstract}
		
		
		As the GAN-based face image and video generation techniques, widely known as DeepFakes, have become more and more matured and realistic, there comes a pressing and urgent demand for effective DeepFakes detectors. Motivated by the fact that remote visual photoplethysmography (PPG) is made possible by monitoring the minuscule periodic changes of skin color due to blood pumping through the face, we conjecture that normal heartbeat rhythms found in the real face videos will be disrupted or even entirely broken in a DeepFake video, making it a potentially powerful indicator for DeepFake detection.
		In this work, we propose \emph{DeepRhythm}, a DeepFake detection technique that exposes DeepFakes by monitoring the heartbeat rhythms. \emph{DeepRhythm} utilizes dual-spatial-temporal attention to adapt to dynamically changing face and fake types. Extensive experiments on FaceForensics++ and DFDC-preview datasets have confirmed our conjecture and demonstrated not only the effectiveness, but also the generalization capability of \emph{DeepRhythm} over different datasets by various DeepFakes generation techniques and multifarious challenging degradations.

	\end{abstract}


	
	\maketitle
	
	\section{Introduction}\label{sec:intro}
	
	\begin{figure}[t]
		\centering
		\includegraphics[width=0.9\linewidth]{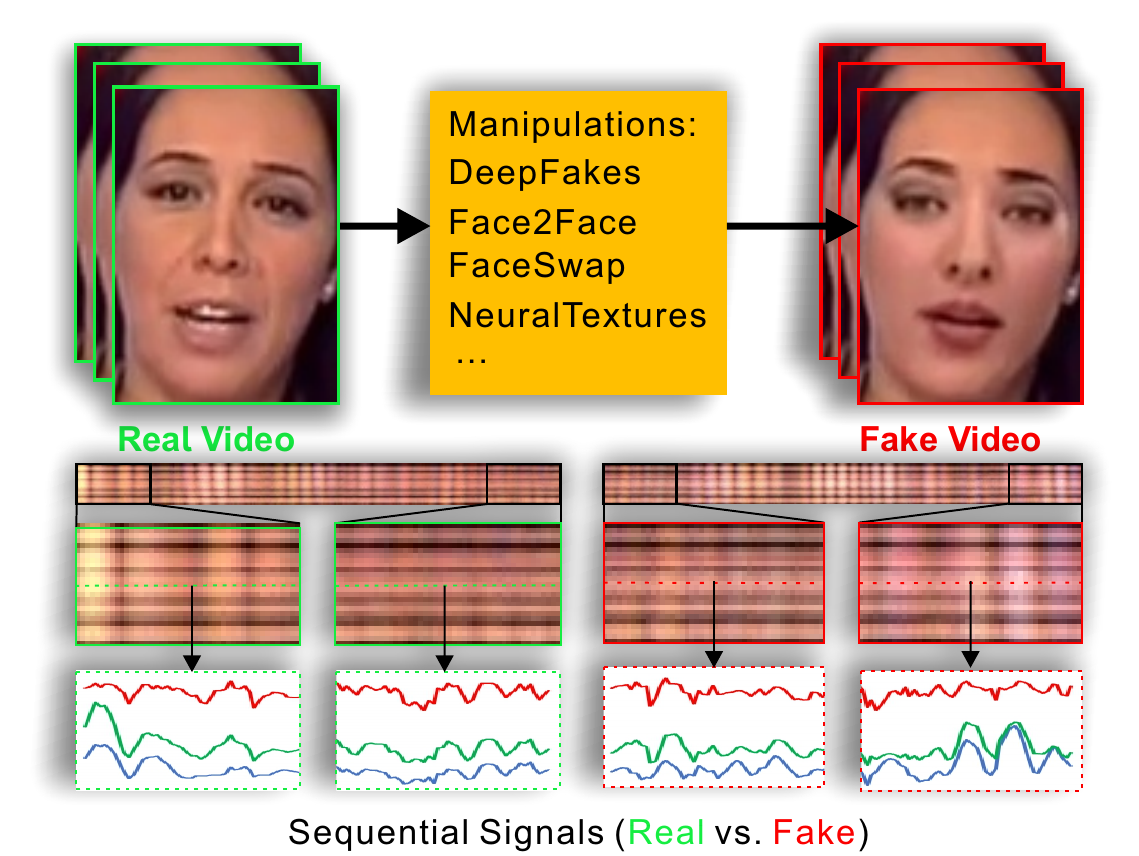}
		\caption{An example of a real video and its fake video generated by various manipulations, \eg, DeepFakes, Face2Face, FaceSwap, \etc \cite{faceforensicspp}. It is hard to decide real/fake via the appearance from a single frame. The state-of-the-art Xception \cite{xception} fails in this case. However, we see that the manipulations easily diminish the sequential signals representing remote heartbeat rhythms.}
		\label{fig:motivation}
	\end{figure}
	
	Over the past decades, multimedia contents such as image and video have become more and more prevalent on various social media platforms. More recently, with the advancement in deep learning-based image and video generation techniques, \ie, generative adversarial networks (GAN) \cite{goodfellow2014gan}, anyone can now generate, \eg, a realistic-looking face that does not exist in the world, or perform a face swap in a video with a high level of realism. The latter is what the community refers to as the DeepFake \cite{karras2017PGGAN,karras2018StyleGAN,karras2019StyleGAN2}. Such a face swap used to require domain expertise such as theatrical visual effects (VFX) and/or high-speed tracking with markers (\eg, motion captures in the movie Avatar). But now, anyone can do it easily. The low barriers to entry and wide accessibility of pre-trained DeepFake generator models are what the problem is. DeepFakes are now a pressing and tangible threat to the integrity of multimedia information available to us. DeepFakes, \eg., when applied on politicians, fueled with targeted misinformation, can really sway people's opinions and can lead to detrimental outcomes such as manipulated and interfered election without people even knowing about it.

	Therefore, the fight against DeepFakes is currently in dire need. Although the detection of DeepFakes is a fairly new research frontier, there have been some attempts to spot DeepFake videos. Some methods are based on traditional digital forensics techniques (see Section \ref{related:forgery}), while others heavily rely on deep learning-based image and video classification \emph{(real vs. fake)} from the raw pixel-domain DeepFake inputs. However, such detection methods based solely on the raw pixel-domain input might become less effective when the DeepFake images and videos become more and more realistic as the deep image generation methods themselves become more mature in the near future. Consequently, a fundamentally different DeepFake detection method is needed. 
	
	In this work, we present \emph{DeepRhythm}, a novel DeepFake detection technique that is intuitively motivated and is designed from ground up with first principles in mind. Motivated by the fact that remote visual photoplethysmography (PPG) \cite{yu2019remote} is made possible by monitoring the minuscule periodic changes of skin color due to blood pumping through the face from a video, we conjecture that normal heartbeat rhythms found in the real face videos will be disrupted or even broken entirely in a DeepFake video, making it a powerful indicator for detecting DeepFakes. As shown in Figure \ref{fig:motivation}, existing manipulations, \eg, DeepFakes, significantly change the sequential signals of the real video, which contains the primary information of the heartbeat rhythm.
	%
	%
	To further make our proposed DeepRhythm method work more robustly under various degradations, we have devised and incorporated both a heart rhythm motion amplification module as well as a learnable spatial-temporal attention mechanism at various stages of the network model. 
	
	Together, through extensive experiments, we have demonstrated that our conjecture holds true and the proposed method indeed effectively exposes DeepFakes by monitoring the heartbeat rhythms. More specifically, DeepRhythm outperforms four state-of-the-art
	DeepFake detection methods including Bayer's method \cite{novelcnn}, Inception ResNet V1 \cite{inception}, Xception \cite{xception}, and MesoNet \cite{mesonet} in the FaceForensics++ benchmark \cite{faceforensicspp} and exhibits high robustness to JPEG compression, noise, and blur degradations.
	To the best of our knowledge, this is the very first attempt to expose DeepFakes using heartbeat rhythms. Our main contributions are summarized as follows:
	\begin{itemize}[leftmargin=*]
		\item We propose \textit{DeepRhythm}, the very first method for effective detection of DeepFake with the heartbeat rhythms.
		
		\item To characterize the sequential signals of face videos, we propose the \textit{motion-magnified spatial-temporal representation (MMSTR)} that provides powerful discriminative features for high accurate DeepFake detection.
		
		\item To fully utilize the MMSTR, we propose \textit{dual-spatial-temporal attention network} to adapt to dynamically changing faces and various fake types. Experimental results on FaceForensics++ and DeepFake Detection Challenge-preview dataset demonstrate that our method not only outperforms state-of-the-art methods but is robust to various degradations.
	\end{itemize}

	
	

	\section{Related Work}\label{sec:related}
	
	\subsection{DeepFakes Generation}
	
	
	Recently, DeepFake techniques have gained widespread attention and been used in generating pornographic videos, fake news, and hoaxes, \etc. Some early studies use face-wrap-based methods to generate fake videos. For example, Bregler \etal \cite{1997videorewrite} track the movement of the speaker's mouth and morph the input video. Dale \etal \cite{video_face_replacement} present a video face replacement via a face 3D model. Similarly, Garrido \etal propose a face warp system while keeping the original face performance \cite{Garrido2014AutomaticFR}  and a photo-realistically replacement method via high-quality monocular capture \cite{VDub}. Thies \etal develop a real-time expression transfer for facial reenactment \cite{Thies2015Realtime} and propose the Face2Face method \cite{Thies2016face2face} that tracks target and source facial expressions to build a face 3D model and re-renders source face on the target model. In addition, Thies \etal \cite{thies2019deferred} further use neural textures and defer neural render to generate the forgeries. 
	
	Besides the above face-wrap-based methods, recent DeepFake approaches, \ie, PGGAN \cite{karras2017PGGAN}, StyleGAN \cite{karras2018StyleGAN}, and StyleGAN2 \cite{karras2019StyleGAN2}, employ the generative adversarial network (GAN) \cite{goodfellow2014gan} for the near-realistic face synthesis. Moreover, some methods can even alter face attributes, \eg, changing or removing the color of the hair, adding glasses or scars \cite{upchurch2016,facial_soft_bio,choi2017stargan,attgan}, and modifying persons' facial expression \cite{liu2019stgan}. Overall, GANs have shown great potential in this area and are easy to use. However, current DeepFake methods, even those based on GANs, do not explicitly preserve the pulse signal, inspiring us to capitalize on the pulse signal to distinguish the real and manipulated videos.
	
	\subsection{Forgery and DeepFake Detection}\label{related:forgery}
	
	DeepFake detection is challenging since GAN-based DeepFakes can generate near-realistic faces that are hardly detected even using the state-of-the-art digital forensics. To alleviate this challenge, researchers are exploring effective solutions to identify fake videos. 
	
	Early attempts focus on detecting forgeries via hand-crafted features, \eg, \cite{imgsplicing,icip16_paint,steganography,3dcooccurrences}. 
	%
	%
	However, these hand-crafted models can be strenuous due to the realistic faces generated by SOTA DeepFake methods (\eg, FaceApp, Reflect, and ZAO). 
	Later, researchers regard the DeepFake detection as a classification problem by extracting discriminative features, \eg, color cues \cite{mccloskey2018colorcue}, monitoring neuron behaviors \cite{ijcai20_fakespotter,deepgauge,deephunter,deepmutation,deepct}, and employing classifiers, \eg, support vector machine (SVM), to tell whether a video is fake or real.
	
	In addition, many researchers also employ SOTA deep neural networks~(DNNs) to detect forgery images. Cozzolino \etal \cite{cnnresidual} use residual-based local features and achieve significant performance. Bayar \etal \cite{novelcnn} and Rahmouni \etal \cite{custompooling} propose novel DNNs to detect manipulated images. Zhou \etal \cite{zhou2018twostream} combine a DNN-based face classification stream with a steganalysis-based triplet stream, yielding good performance. More recently, researchers are trying to apply much more complex and advanced DNNs on video forgery detection, such as Inception-ResNet \cite{inception}, MesoNet \cite{mesonet}, capsule networks \cite{nguyen2019capnet}, and Xception \cite{xception}. 
	
	Besides only adopting convolutional neural networks (CNNs), some researchers use a combined recurrent neural network (RNN) and CNN to extract image and temporal features to distinguish real and fake videos. For example, Güera \etal \cite{guera2018cnnrnn} use CNN features to train an RNN to classify videos. Similarly, Sabir \etal \cite{sabir2019recurrent} use a sequence of spatio-temporal faces, as RNN's input to classify the videos. Furthermore, Dong \etal \cite{dang2019detection} utilize an attention mechanism to generate and improve the feature maps, which highlight the informative regions. 
	Different from existing methods, our technique initiates the first step of leveraging remote heartbeat rhythms for DeepFake detection. To achieve high detection accuracy, we propose a motion-magnified representation for the heartbeat rhythms and employ a spatial-temporal representation to improve the ability in distinguishing real and fake videos.
	\subsection{Remote Photoplethysmography (rPPG) Anti-Spoofing} 
	
	
	
	
	
	
	Face spoof detection is similar to DeepFake detection, aiming to determine whether a video contains a live face.
	Since the remote heart rhythm~(HR) measuring techniques achieve quite a bit of progress \cite{2010hrmonitor,2007dualppg,poh2010pulsemeasure,poh2011pulsemeasure,balak2013motionpulse,li2014remotehr,yu2019remote}, many works use rPPG for face spoofing detection. For example, Li \etal \cite{pulsemask} use the pulse difference between real and printed faces to defend spoofing attacks. Nowara \etal \cite{ppgbg} compare PPGs of face and background to decide whether the face is live or not. Heusch \etal \cite{ppgltss} use the long-term statistical spectral on the pulse signals and Hernandez-Ortega \etal \cite{ppgnir} employ the near infrared against realistic artifacts. Moreover, combining with DNNs, Liu \etal \cite{ppgsupervision} extract spatial and temporal auxiliary information, \eg, depth map and rPPG signal, to distinguish whether it is a live face or spoofing face. 
	

	Overall, existing anti-spoofing methods also benefit from employing rPPG for liveness detection, which seems similar to our work. However, there are fundamental differences: the liveness detection mainly relies on the judgment about whether the heart rhythms exist or not; our work aims to find the different patterns between real and fake heart rhythms since fake videos may still have the heart rhythms but their patterns are diminished by DeepFake methods and are different from the real ones (see Figure~\ref{fig:motivation}).


	
	
	\section{Method}\label{sec:method}
\begin{figure*}
		\centering
		\includegraphics[width=0.95\linewidth]{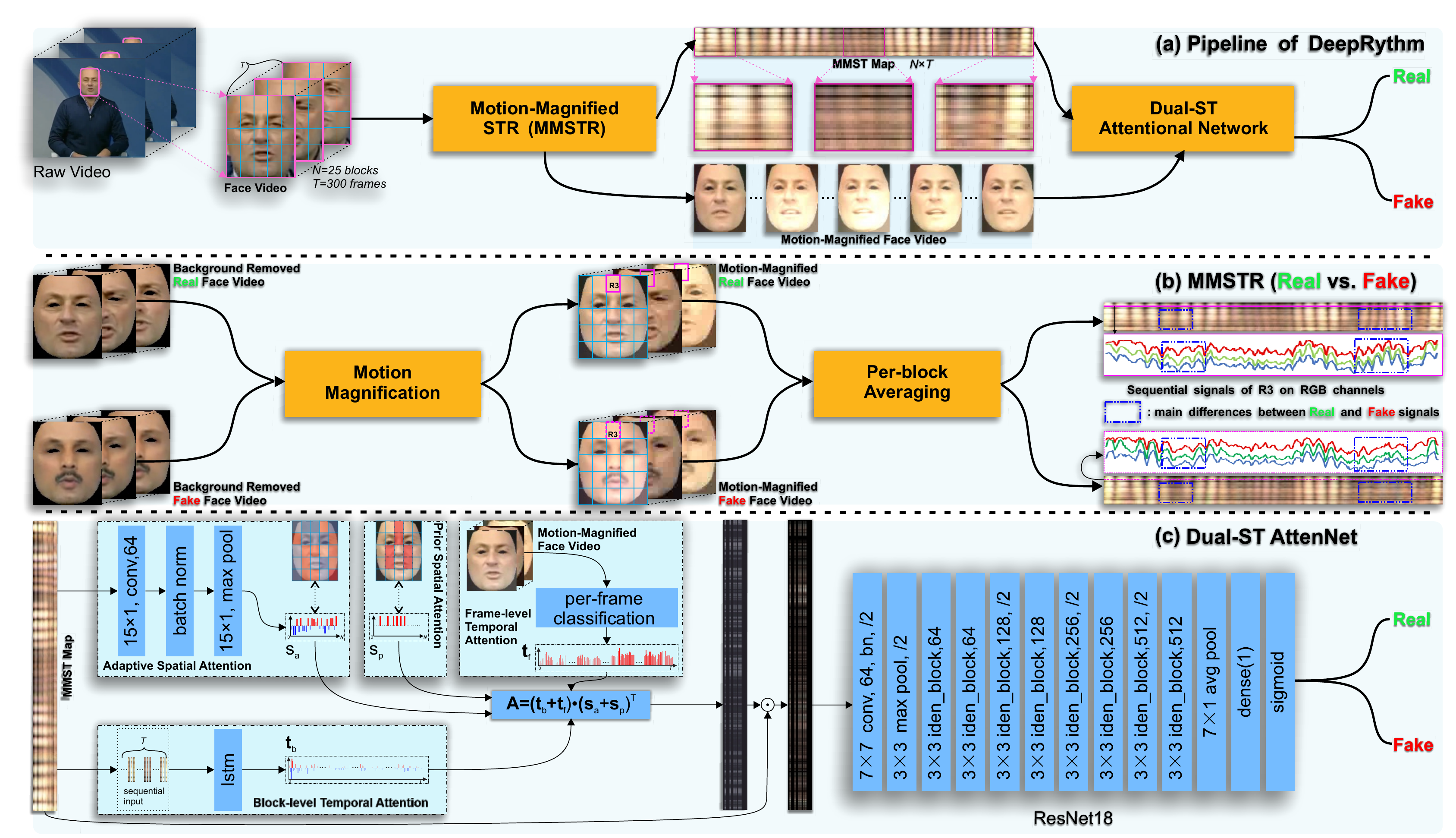}
		\caption{The workflow of DeepRhythm, \ie, (a), and its two main modules: motion-magnified spatial-temporal representation (MMSTR), \ie, (b), and dual-spatial-temporal attentional network~(Dual-ST AttenNet), \ie, (c). We also highlight the main differences of MMSTRs between real and fake videos in sub-figure (b).}
		\label{fig:pipeline}
	\end{figure*}

	We propose \textit{DeepRhythm}~(Sec.~\ref{subsec:deeprhythm}) for effective DeepFake detection by judging whether the normal HR in face videos are diminished. Figure~\ref{fig:pipeline}~(a) summarizes the workflow of DeepRhythm. 
	%
	\subsection{DeepRhythm for DeepFake Detection}
	\label{subsec:deeprhythm}
	Given a face video $\mathcal{V}=\{\mathbf{I}_i\}_{i=1}^{T}$ that contains $T$ frames, our goal is to predict if this video is real or fake according to the heart rhythm signals. To this end, we first develop the \textit{motion-magnified spatial-temporal representation (MMSTR)~(Sec.~\ref{subsec:mmstr})} for face videos, which can highlight the heart rhythm signals and output a \textit{motion-magnified spatial-temporal map (MMST map)}, \ie, $\mathbf{X}=\mathrm{mmstr}(\mathcal{V})\in\mathds{R}^{T\times N \times C}$ where $T$ is the number of frames, $N$ is the $N$ 
	region of interest (ROI) blocks of the face in $\mathcal{V}$~(\ie, the regions marked by the blur grid in Figure~\ref{fig:pipeline}~(a)), and $C$ means the number of color channels. In the following, we formulate with single color channel for clear representation but use RGB channels in practice. Intuitively, $\mathbf{X}$ contains the motion-magnified temporal variation of $N$ blocks in the face video, \ie, highlighted heart rhythm signals.

	We can simply design a deep neural network that takes the MMST map as input and predict if the raw video is real. However, various interference, \eg, head movement, illumination variation, and sensor noises, may corrupt the MMST map. As a result, the contributions of different positions in the MMST map are not always the same (\eg, the three patches shown in Figure~\ref{fig:pipeline}~(a) have different heart rhythm strength), which definitely affects the fake detection accuracy. To alleviate this challenge, we should assign different weights to different positions of the MMST map before further performing the fake detection
	\begin{align}\label{eq:rhythmv1}
	\mathbf{y}=\phi(\mathbf{A}\odot \mathbf{X}),
	\end{align}
	where $\phi(\cdot)$ is a CNN for real/fake classification, $\odot$ denotes the element-wise multiplication, and $\mathbf{y}$ is the prediction (\ie, 1 for fake and 0 for real). The matrix $\mathbf{A}\in \mathds{R}^{T\times N}$ provides different weights to different positions of $\mathbf{X}$ and is known as an attention mechanism. We let RGB channels share the attention matrix.
	
	We aim to produce $\mathbf{A}$ via a DNN. However, due to the diverse types of fake and dynamic changing faces, it is difficult to get proper $\mathbf{A}$ for different face directly. We handle this problem by further decomposing $\mathbf{A}$ into two parts, \ie, spatial attention $\mathbf{s}\in\mathds{R}^{N\times 1}$ and temporal attention $\mathbf{t}\in\mathds{R}^{T\times 1}$, and reformulate Eq.~(\ref{eq:rhythmv1}) as
	\begin{align}\label{eq:rhythmv2}
	\mathbf{y}=\phi((\mathbf{t}\cdot\mathbf{s}^{\top})\odot \mathbf{X}),
	\end{align}
	Intuitively, the two attentions indicate when~(along the $T$'s axis) and where~(along the $N$'s axis) of the input MMST map should be used for better fake detection. Furthermore, the number of parameters of $\mathbf{s}$ and $\mathbf{t}$, \ie, $N+T$, is much smaller than that of $\mathbf{A}$, \ie, $N\cdot T$, which allows the spatial-temporal-attention to be tuned more easily.
	
	Then, the key problem is how to generate $\mathbf{t}$ and $\mathbf{s}$ to adapt to dynamically changing faces and various fake types. In Sec.~\ref{subsec:dstatten}, we propose the dual-spatial-temporal attention network to realize Eq.~(\ref{eq:rhythmv2}) by jointly considering prior~\&~adaptive spatial attention and frame~\&~block temporal attention.

	\subsection{Motion-Magnified Spatial-Temporal Representation}
	\label{subsec:mmstr}
	A straightforward way of employing heart rate (HR) signals for DeepFake detection is to use existing HR representations that are designed for the remote HR estimation. For example, we can use the spatial-temporal representation (STR) proposed by Niu \etal \cite{Niu2020TIP} for representing HR signals and feed them to a classifier for DeepFake detection. However, it is hard to achieve high fake detection accuracy with the STR directly since the differences between real and fake videos are not highlighted, \ie, STR's discriminative power for DeepFake detection is limited.
	
	To alleviate the problem, we propose the \textit{motion-magnified STR (MMSTR)} where differences between real and fake face videos can be effectively represented. Specifically, given in a face video, \ie, $\mathcal{V}$ having $T$ frames, we calculate MMSTR using the following steps:
	\begin{enumerate}[label=(\roman*)]
		\item Calculate landmarks\footnote{ \url{https://github.com/codeniko/shape_predictor_81_face_landmarks}.} of the faces on all frames of $\mathcal{V}$ and remove the eyes and background according to the landmarks, \eg, the faces shown in the left of Figure~\ref{fig:pipeline}~(b).
		\item Perform the motion magnification algorithm~\cite{Wu12Eulerian,oh2018learningbased}\footnote{We use its python implementation: \url{https://github.com/flyingzhao/PyEVM}.} on the background removed face video and obtain motion-magnified face video with RGB space.
		\item Divide the face areas of all frames into $N$ non-overlapping ROI blocks, \ie, regions marked by the blue grid in Figure~\ref{fig:pipeline}~(b), and perform average pooling on each block and each color channel for each frame. We then obtain the MMST map, \ie, $\mathbf{X}$, as the sub-figures shown in the right of Figure~\ref{fig:pipeline}~(b). Each row of $\mathbf{X}$ represents the motion-magnified temporal variation of one block on RGB channels, as the red, green, and blue curves shown in Figure~\ref{fig:pipeline}~(b). 
	\end{enumerate}
	
	Figure~\ref{fig:pipeline}~(b) shows examples of real and fake face videos and their MMST maps, respectively. We have the following observations: 1) it is difficult to judge which video is fake just by looking at the raw frames. 2) differences between the real and fake videos can be easily found on our MMST maps that will provide effective information for fake detection. The advantages of MMSTR over STR will be further discussed in the experimental section.
	
	\subsection{Dual-Spatial-Temporal Attentional Network}
	\label{subsec:dstatten}
	In this section, we detail the dual-spatial-temporal attentional network (Dual-ST AttenNet), with which we can realize accurate DeepFake detection through the MMST map and its spatial and temporal attentions, \ie, $\mathbf{t}$ and $\mathbf{s}$ defined in Eq.~(\ref{eq:rhythmv2}).
	
	\subsubsection{Dual-Spatial Attention}\label{subsubsec:s_atten}
	
	The dual-spatial attention is
	\begin{align}\label{eq:sp-atten}
	\mathbf{s}=\mathbf{s}_\mathrm{a}+\mathbf{s}_\mathrm{p},
	\end{align}
	where $\mathbf{s}_\mathrm{p}\in\mathds{R}^{N\times 1}$ and $\mathbf{s}_\mathrm{a}\in\mathds{R}^{N\times 1}$ are the prior and face-adaptive spatial attentions, respectively. The prior attention $\mathbf{s}_\mathrm{p}$ is a fixed vector whose six specified elements are set as one while others are zero, which is to extract the HR signals from six specified ROI blocks and ignore signals from other blocks. 
	The six specified ROI blocks are the four blocks under eyes and two blocks between eyes, as shown in Figure~\ref{fig:pipeline}~(c). The intuition behind this idea is that the specified blocks are usually robust to various real-world interference while the HR signals of other blocks are easily diminished when unexpected situations happen, \eg, head movement might let the HR signals of blocks at face boundary disappear.
	
	In addition to the prior spatial attention, we also need face-adaptive attention, \ie, $\mathbf{s}_\mathrm{a}$, to highlight different blocks to adapt to the environment variations since even the same face under different situations, \eg, the illumination changes, has different effective ROI blocks. To this end, we propose to train a spatial attention network to generate adaptive spatial attention, which contains a convolution layer that has 64 kernels with size being $15\times 1$ followed by a batch normalization layer and max-pooling layer. The CNN's parameters are jointly learned with the whole framework.
	
	\subsubsection{Dual-Temporal Attention}
	\label{subsubsec:t_atten}
	
	DeepFake methods usually add different fake textures at different face locations to different frames, which not only destroy the smooth temporal variation of a face but lead to inconsistent fake magnitude among frames (\ie, some frames contain obvious fake textures while others have few or no fakes). We propose dual-temporal attention to consider above information
	\begin{align}\label{eq:tm-atten}
	\mathbf{t}=\mathbf{t}_\mathrm{b}+\mathbf{t}_\mathrm{f},
	\end{align}
	where $\mathbf{t}_\mathrm{b}\in\mathds{R}^{T\times 1}$ and $\mathbf{t}_\mathrm{f}\in\mathds{R}^{T\times 1}$ indicate which frames are more significant for final fake detection. Specifically, we train an LSTM to represent the temporal variation of a face, which is sequentially fed with each row of the MMST map, \ie, $\mathbf{X}$, and outputs $\mathbf{t}_\mathrm{b}$ that is denoted as the block-level temporal attention. The LSTM's parameters are jointly trained with the whole framework.
	
	To take full advantage of fake textures in each frame, we train a temporal-attention network that takes each motion-magnified frame as input and scores the fakeness of the frame independently and get $\mathbf{t}_\mathrm{f}$. The frames with higher probability to be fake contribute more to the final classification and we denote $\mathbf{t}_\mathrm{f}$ as the frame-level temporal attention. In practice, we use the Meso-4 architecture \cite{mesonet} as the network containing a sequence of four convolution layers and two fully-connected layers. The Meso-4's parameters are independently trained for frame-level fake detection.
	
	\subsubsection{Implementation details}
	\label{subsubsec:impl_details}
	
	Our dual-spatial-temporal attentional network is shown in Figure~\ref{fig:pipeline}~(c) where an MMST map, \ie, $\mathbf{X}$, is first employed to produce the adaptive spatial attention, \ie, $\mathbf{s}_\mathrm{a}$ and the block-level temporal attention, \ie, $\mathbf{t}_\mathrm{b}$, through a spatial attention network and an LSTM, respectively. The pre-trained Meso-4 is fed with the motion-magnified face video and outputs the frame-level temporal attention, \ie, $\mathbf{t}_\mathrm{f}$. Finally, the attentional MMST map, \ie, $(\mathbf{t}\cdot\mathbf{s}^{\top})\cdot \mathbf{X}$, is fed to the $\phi(\cdot)$ for the final DeepFake detection where we use ResNet18 \cite{He2016CVPR} for the $\phi(\cdot)$.
	
	We jointly train parameters of the spatial attention network, the LSTM, and the ResNet18 using the cross-entropy loss with Adam optimizer. The learning rate and weight decay are set as 0.1 and 0.01 respectively. The max epoch number is set to 500, and training will stop if validation loss did not decrease in 50 epochs. For training the Meso-4, we use the same hyper-parameters. We use videos from FaceForensics++ \cite{faceforensicspp} as the training dataset which is introduced in Sec.~\ref{subsec:dataset} and Table~\ref{tab:datasets}. Our implementation and results are obtained on a server with Intel Xeon E5-1650-v4 CPU and NVIDIA GP102L.

	\section{Experiments}\label{sec:exp}
	\begin{figure}
		\centering
		\includegraphics[width=0.9\linewidth]{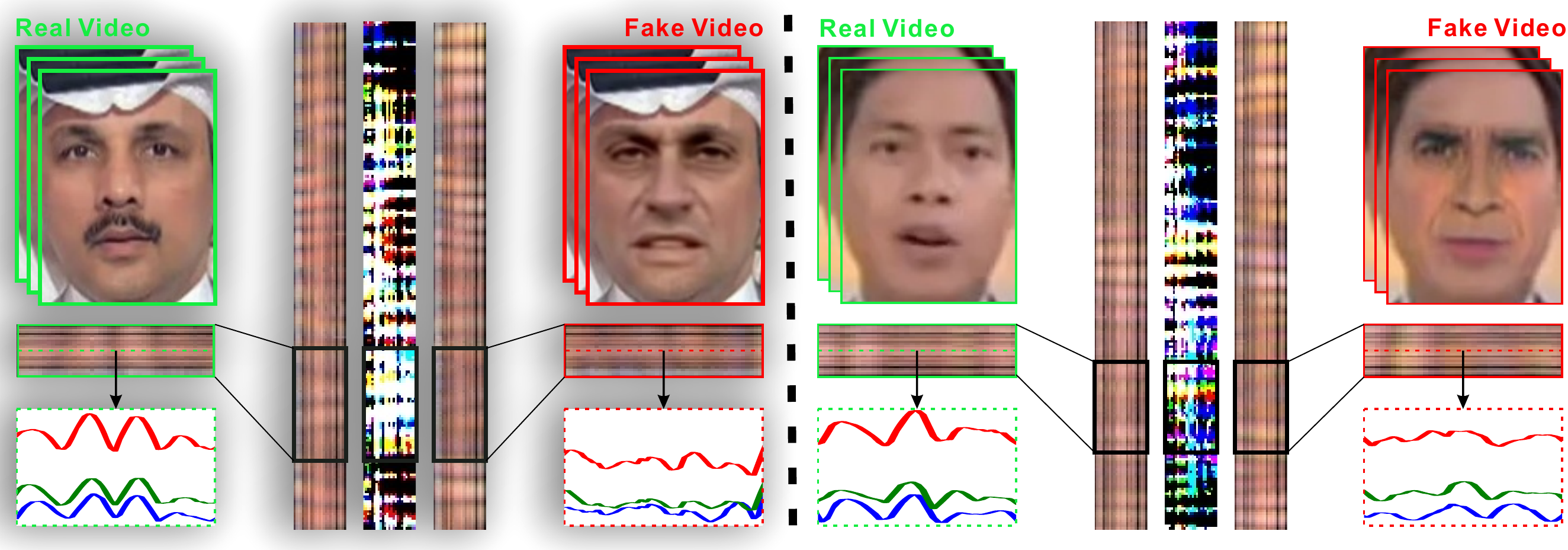} 
		\caption{Two real-fake video pairs, their MMST maps and the colorful difference maps between real and fake. 
		}  
		\label{fig:vis_baselines}
	\end{figure}
	
	
	\subsection{Dataset and Experiment Setting}\label{subsec:ex_dataset}
	\label{subsec:dataset}
	
	{\bf Dataset.} 
	We select FaceForensics++ \cite{faceforensicspp} as our training and testing datasets, and use DeepFake Detection Challenge preview (DFDC-preview) \cite{dfdc_preview} as an additional testing dataset to evaluate our method's cross-dataset generalization capability. 
	
	FaceForensics++ dataset consists of thousands of videos manipulated with different DeepFake methods and contains four fake sub-datasets, \ie, DeepFake Detection (DFD), DeepFake (DF), Face2Face (F2F), and FaceSwap (FS). However, the original FaceForensics++ dataset has the data imbalance problem. For example, the original DFD subset contains 2728 fake videos, but only 268 real videos. To solve this problem, we make the following improvements: 1) we augment the 
	the original 268 real videos by flipping them horizontally and get a total of 2,510 real videos. 2) To evaluate our method on the whole FaceForensics++ dataset, we build an extra dataset, \ie `ALL' in Table~\ref{tab:datasets}, by concatenating the four subsets and augment the real videos by flipping them horizontally and vertically, and rotating them 180 degrees, respectively. Table \ref{tab:datasets} summarizes the final four subsets, the ALL dataset, and their partitions about training, validation, and testing. The dataset partition ratio is 8:1:1 and the augmented videos are removed in the testing datasets.
	
	
	
	For videos in DeepFake Detection Challenge, we directly use it as the testing set. The details can be found in Table~\ref{tab:datasets}. Directly using the DFDC-preview dataset as testing set will cause imbalance between the number of real and fake videos. So we randomly sample 500 videos from the real part and 500 videos from the fake part, and assemble them as the testing set. Then, we test every method formerly trained on ALL's training set on it to compare their performance.
	
	{\bf Pre-processing.}
	%
	For every video in the testing and training datasets, we take the first 300 frames to produce the MMST map. Specifically, while processing frames, we first use MTCNN \cite{facenet_pytorch} to detect face, and then use Dlib to get 81 facial landmarks \cite{81facelandmarks}. If faces are not detected in a frame, this frame will be abandoned. If more than 50 frames were abandoned, this video will not be used to train the network. If more than one faces were detected in one frame, the one closer to the faces of previous frames will be retained.

	{\bf Baseline.}
	We choose the state-of-the-art DeepFake detection methods, \ie, Bayer's method \cite{novelcnn}, Inception ResNet V1 \cite{inception}, Xception \cite{xception} and MesoNet \cite{mesonet}, as baselines. All of them obtain high performance on FaceForensics++'s benchmark \cite{benchmark} and are the top-4 accessible methods therewithin. For Bayer's method \cite{novelcnn}, we do not find publicly available code, so we re-implement it on Keras. For Inception ResNet V1 \cite{inception} and Xception \cite{xception}, we directly use the Keras code provided and only add a Dense layer with one neuron after the final layer to get prediction. As for MesoNet \cite{mesonet}, we directly use the code provided by authors \cite{MesoNet_git}. 
	
	It should be noted that these baselines perform the fake detection on an image instead of a video, \ie, estimating if a frame is real or fake. We make the following adaptations to make them suitable to address videos: 1) For the testing setup, we use these baselines to predict every frame of the video and count the number of real or fake frames. If the real frames are more than fake ones, we identified this video as real, and vice versa. 2) In terms of the training setup, we take the first five frames for every training video in the `ALL' dataset in Table~\ref{tab:datasets} and extract their facial region via MTCNN, all of these faces are divided into training, validation, and testing subsets. We also employ Adam optimizer with batch-size of 32 and learning rate of 0.001. The max epoch number is 500, and the training will stop if validation loss did not decrease in 50 epochs.

	\begin{table}
		\centering
		\caption{Details of FaceForensics++~(FF++) and DFDC-preview datasets for both testing and training}
		\label{tab:datasets}
		\begin{tabular}{c|c|c|c|c||c|c|c}
			\hline
			\rowcolor{tabgray2} \multicolumn{2}{c|}{Dataset} & total & real & fake & train & val. & test \\
			\hline 
			\multirow{5}{*}{FF++}& DFD & 5238 & 2510 & 2728 & 4190 & 524 & 524 \\
			& DF & 1959 & 988 & 971 & 1567 & 195 & 197 \\
			& F2F & 1966 & 988 & 978 & 1572 & 196 & 198 \\
			& FS & 1971 & 988 & 983 & 1582 & 197 & 198 \\
			& ALL & 10680 & 5020 & 5660 & 8544 & 1068 & 1068 \\
			\hline 
			\multicolumn{2}{c|}{DFDC-preview} & 3310 & 578 & 2732 & - & - & 1000 \\
			\hline 
		\end{tabular}
	\end{table}

	\begin{table*}
		\centering
		\caption{Comparison with baseline methods on FaceForensics++ and DFDC-preview datasets with the models trained on sub-datasets and ALL dataset of FaceForensics++, respectively. We highlight the best and second best results with red and yellow.}
		\label{tab:comparison_baseline}
		\begin{tabular}{c|c|c|c|c|c|c|c|c|c|c}
			\hline
			\rowcolor{tabgray2} & \multicolumn{4}{c}{train on sub-datasets} & \multicolumn{6}{|c}{train on ALL dataset} \\
			\hline 
			\rowcolor{tabgray2} test on & DFD & DF & F2F & FS & DFD & DF & F2F & FS & ALL & DFDC \\
			\hline 
			Bayer and Stamm \cite{novelcnn}     & 0.52  & 0.503 & 0.505 & 0.505 & 0.501 & 0.52 & 0.503 & 0.505 & 0.5 & 0.5 \\
			Inception ResNet V1 \cite{inception} & 0.794 & 0.783 & 0.788 & 0.778 & 0.919 & 0.638 & 0.566 & 0.462 & 0.774 & 0.597 \\
			Xception \cite{xception}            & \cellcolor{top2} 0.98  & \cellcolor{top2} 0.995 & \cellcolor{top2} 0.985 & 0.98  & \cellcolor{top2} 0.965 & \cellcolor{top2} 0.984 & \cellcolor{top2} 0.984  &  \cellcolor{top2} 0.97 & \cellcolor{top2} 0.978 & 0.612 \\
			MesoNet \cite{mesonet}             & 0.804 & 0.979 & \cellcolor{top2} 0.985 & \cellcolor{top2} 0.995 & 0.958 & 0.822 & 0.813 & 0.783 & 0.909 &\cellcolor{top1} 0.745 \\
			\textbf{DeepRhythm (ours)}         & \cellcolor{top1} 0.987 & \cellcolor{top1} 1.0   & \cellcolor{top1} 0.995 & \cellcolor{top1} 1.0   & \cellcolor{top1} 0.975 & \cellcolor{top1} 0.997 & \cellcolor{top1} 0.989 & \cellcolor{top1} 0.978  & \cellcolor{top1} 0.98  & \cellcolor{top2} 0.641 \\
			\hline 
		\end{tabular}
	\end{table*}
	
	\subsection{Baseline Comparison on Accuracy}\label{subsec:ex_baseline}

	We test all methods on DFD, DF, F2F, FS, and ALL subsets (reported in Table~\ref{tab:comparison_baseline}) and the DFDC-preview with the models trained on FF++'s subsets (DFD, DF, F2F, FS, and ALL), respectively. 
	
	{\bf Results on FaceForensics++.}
	Overall, in Table~\ref{tab:comparison_baseline}, our DeepRhythm achieves the highest accuracy on all datasets compared with the baseline methods.
	First, our method gets better results than other methods across all cases, regardless of which training dataset is used. This demonstrates the generalization capability of our method across various DeepFake techniques. 
	Second, although we adopt the MesoNet in our framework for the frame-level temporal attention, our method significantly outperforms the MesoNet on all cases, \eg, when trained on ALL dataset, DeepRhythm achieves 0.96 on the FS while MesoNet only has 0.719, which validates the effectiveness of our MMST representation and other attention information, and also indicates the potential capability of our framework for enhancing existing frame-level DeepFake detection methods.
	Third, although the baseline method Xception has obtained significantly high accuracy, \eg, 0.985 and 0.995 on the testing dataset of F2F and FS, it is still exceeded by our method, confirming the advantage of our method over the state-of-the-arts.
	We show two cases from the FaceForensics++ dataset in Figure~\ref{fig:vis_baselines}, where all baseline methods fail to recognize the fake videos while our method succeeds. The fake techniques diminish the sequential signal patterns of real videos (\eg, the waveform of the real video in the first case becomes flat in the fake video), which are effectively captured by our MMST maps.

	{\bf Results on DFDC-preview.}
	According to the results on the DFDC (see Table~\ref{tab:comparison_baseline}), Bayer's still performs worse than others, achieving 0.5 accuracy. Inception ResNet V1 has worse performance than their results on ALL's testing set, achieving 0.597. \textcolor{black}{Xception obtains 0.612 accuracy on DFDC-preview, and is also much worse than its performance on ALL's testing set. Our DeepRhythm gets 0.641 accuracy and is better than Xception, being the second highest accuracy.} Although MesoNet performs not as good on ALL's testing set, it achieves the highest accuracy (\ie, 0.745).

	\begin{figure}[]
		\centering
		\includegraphics[width=0.9\linewidth]{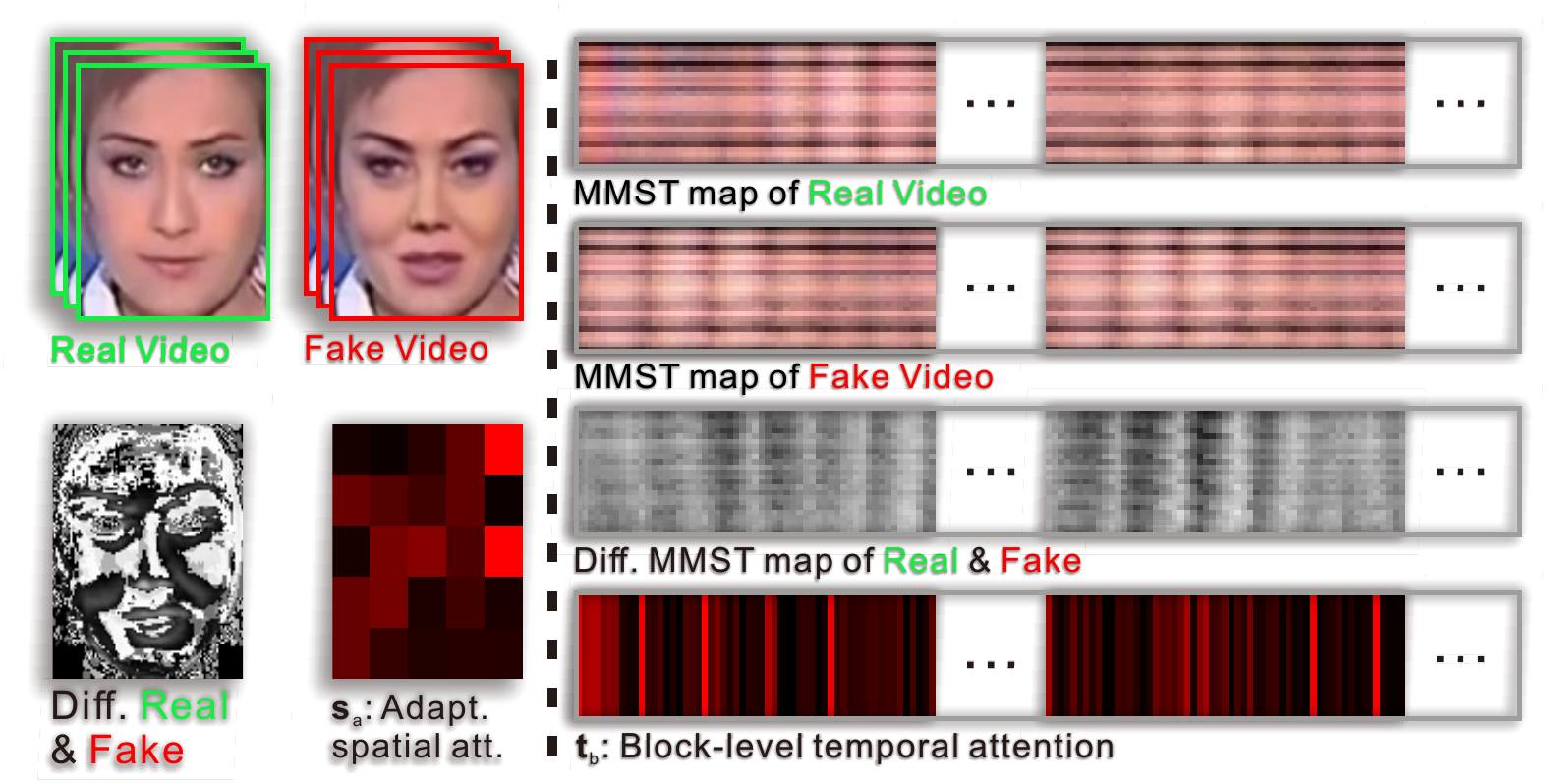} 
		\caption{An example of a real video, the corresponding fake video, the difference image between real and fake frames~(Diff.~Real~\&~Fake), the MMST maps of real and fake videos, the difference map between real and fake MMST maps~(Diff.~MMST map of Real~\&~Fake), the adaptive spatial attention~($\mathbf{s}_\mathrm{a}$, \ie, adapt. spatial att.), and the block-level temporal attention~($\mathbf{t}_\mathrm{b}$).}
		\label{fig:vis_att}
	\end{figure}
	
	\subsection{Ablation Study on Accuracy}\label{subsec:ex_ablation}
	
	To demonstrate the effectiveness of our motion-magnified spatial-temporal representation (MMSTR), dual-spatial-temporal attention network, and end-to-end training, we conduct an ablation study by first training the basic model with existing spatial-temporal (ST) map at the beginning and then add our contributions one by one. 
	
	{\bf DeepRhythm variants.} 
	We first train the bare model (DR-st), which only uses ST map as its input without motion magnification and attention. Then, we use \textit{MMST map} as inputs and re-train our model~(denoted as DR-mmst), still not using any attention. After that, based on the pre-trained DR-mmst, we add \textit{adaptive spatial attention}~(A) and \textit{block-level temporal attention} (B), respectively, (\ie, DR-mmst-A and DR-mmst-B) and perform fine-tuning. After 50 epochs, we observe that the validation loss does not decrease further. Then, we add \textit{prior spatial attention}~(P) and \textit{frame-level temporal attention}~(F) on DR-mmst-A and DR-mmst-B, respectively, then get DR-mmst-AP and DR-mmst-BF that are further fine-tuned. Next, based on either DR-mmst-AP or DR-mmst-BF, we use four attentions together (\ie, DR-mmst-APBF) and carry on training. Finally, we compare DR-mmst-APBF with our final version (\ie, DR-mmst-APBF-e2e), where the \textit{adaptive spatial attention} (A), \textit{block-level temporal attention}~(B), and the network are jointly or end-to-end trained. 
	For all the experiments, we use the same hyper-parameters and datasets, as introduced in Sec.~\ref{subsubsec:impl_details}. The results are summarized in Table~\ref{tab:ablation study}.
	
	{\bf Effectiveness of MMSTR.}
	As shown in Table~\ref{tab:ablation study}, our MMSTR significantly improves the DR-st's accuracy, \eg, 0.328 improvement on ALL. The ST map from \cite{Niu2020TIP} has little discriminative power for DeepFake detection since DR-st achieves about 0.5 accuracy on every testing dataset, which means it randomly guesses a video being real/fake. After using our MMSTR, DR-mmst achieves 0.217 averaged accuracy increment over DFD, DF, F2F, FS, and ALL datasets.
	
	{\bf Effectiveness of single attention.}
	Based on the DR-mmst, we add adaptive spatial attention (DR-mmst-A) and block-level temporal attention (DR-mmst-B), respectively. These two attentions do help improve the model's accuracy, as presented in Table~\ref{tab:ablation study} where DR-mmst-A and DR-mmst-B get average 0.061 and 0.0632 improvements over DR-mmst, respectively.
	
	We further show an example of the adaptive spatial attention~($\mathbf{s}_\mathrm{a}$) and the block-level temporal attention~($\mathbf{t}_\mathrm{b}$) in Figure~\ref{fig:vis_att}. To validate their effectiveness, we also present the difference image and MMST map between real and fake cases. From the view of spatial domain, the difference image indicates that the main changes caused by the fake is around the nose, which is identical to the estimated adaptive spatial attention. In terms of the temporal domain, the estimated temporal attention has high values at the peaks of the difference MMST map.
	
	{\bf Effectiveness of dual-spatial attention.}
	In addition to the adaptive spatial attention~(DR-mmst-A), we further consider the prior attention where the specified ROI blocks on faces are considered and realize the DR-mmst-AP. As validated in Table~\ref{tab:ablation study}, DR-mmst-AP outperforms DR-mmst-A on all compared datasets and obtains an average of 0.033 improvement, which demonstrates the advantage of dual-spatial attention over single adaptive spatial attention.
	
	{\bf Effectiveness of dual-temporal attention.}
	The block-level temporal attention misses details among frames. To alleviate this issue, we add the frame-level temporal attention~(F) to DR-mmst-B for the frame-level DeepFake detection and get the DR-mmst-BF. In Table~\ref{tab:ablation study}, DR-mmst-BF has much higher accuracy than DR-mmst-B on all compared datasets. The average improvement is 0.178, which shows the effectiveness of our dual-temporal attention.
	
	{\bf Effectiveness of dual-spatial-temporal attention and end-to-end training.}
	We put DR-mmst-AP and DR-mmst-BF together and get DR-mmst-APBF. Compared with DR-mmst-AP, DR-mmst-APBF has much higher accuracy on all datasets. However, when comparing it with DR-mmst-BF, DR-mmst-APBF's accuracy slightly decreases on DFD, DF, and ALL while increasing on FS dataset. Though, when we train DR-mmst-APBF in the end-to-end way and get DR-mmst-SPTM-e2e, it achieves the highest accuracy on all testing datasets, indicating that training four attention separately might not mine the potential power of the four attention effectively, and training them together helps get the maximum effect.

	\begin{figure}[t]
		\centering
		\includegraphics[width=0.85\linewidth]{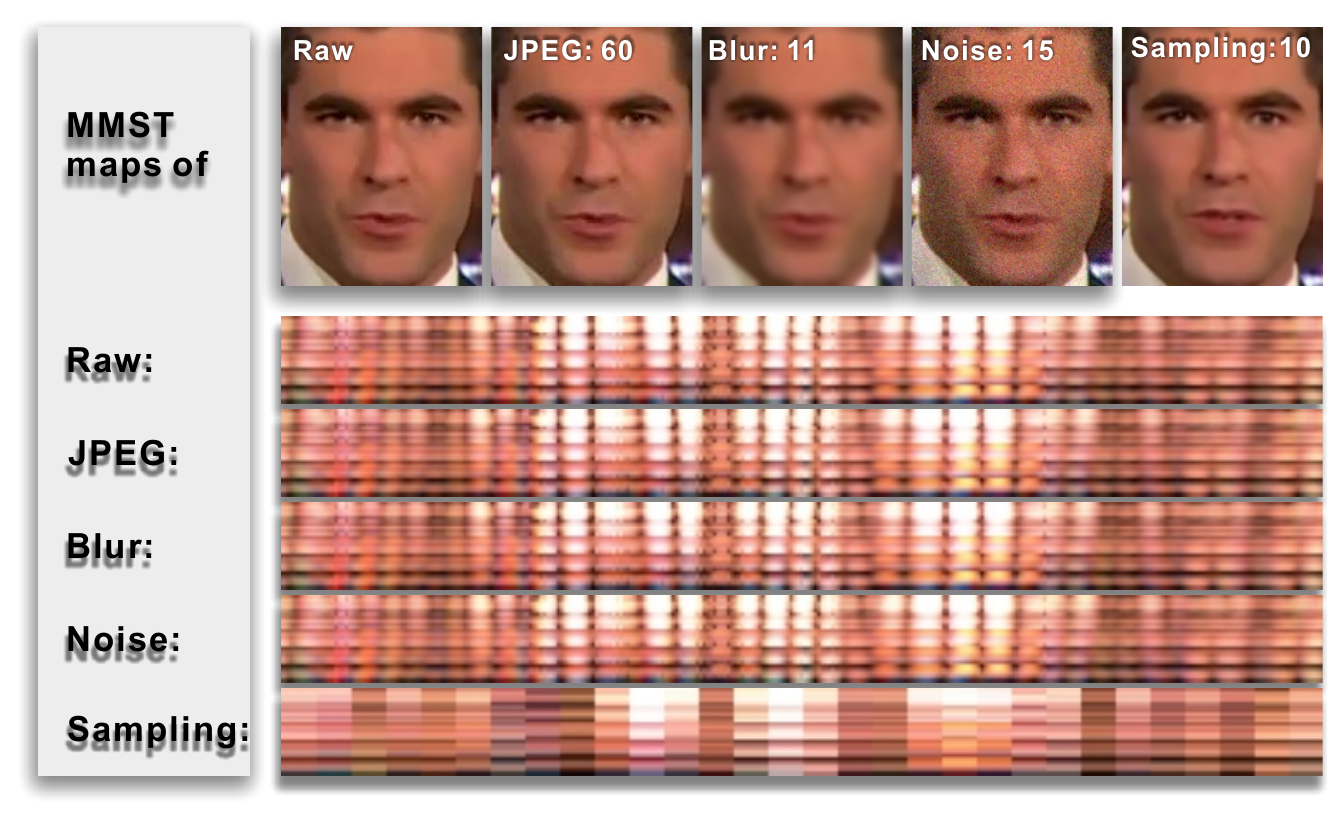}
		\caption{Frames and MMST maps of a video and their four degraded versions with JPEG, blur, noise, and temporal sampling degradations whose degrees are 60, 11, 15, and 10, respectively, which are the median values in the x-axis ranges in Figure~\ref{fig:rob_abla}. Clearly, the JPEG, blur, and noise degradations do not affect the MMST maps of raw videos. The temporal sampling significantly diminishes the raw pattern of the MMST maps. 
		}
		\label{fig:degrads}
	\end{figure}
	
	\begin{table}
		\centering
		\caption{Ablation study of DeepRhythm~(DR) by progressively adding the MMSTR, adaptive~(A) and prior~(P) spatial attentions, block-level~(B) and frame-level~(F) temporal attentions, and end-to-end~(e2e) training strategy.}
		\label{tab:ablation study}
		\begin{tabular}{l|c|c|c|c|c}
			\hline
			\rowcolor{tabgray2} & \multicolumn{5}{c}{train on ALL sub-dataset} \\
			\hline 
			\rowcolor{tabgray2} test on & DFD & DF & F2F & FS & ALL \\
			\hline 
			DR-st               & 0.522 & 0.497 & 0.497 & 0.492 & 0.512 \\
			\rowcolor{top1-7} DR-mmst             & 0.814 & 0.684 & 0.635 & 0.64  & 0.84  \\ \hline
			\rowcolor{top1-6} DR-mmst-A           & 0.849 & 0.77  & 0.736 & 0.716 & 0.847 \\
			\rowcolor{top1-5} DR-mmst-B           & 0.872 & 0.745 & 0.731 & 0.731 & 0.85  \\
			\rowcolor{top1-4} DR-mmst-AP          & 0.879 & 0.816 & 0.766 & 0.756 & 0.867 \\
			\rowcolor{top1-3} DR-mmst-BF          & 0.97  & 0.969 & 0.954 & 0.959 & 0.966 \\
			\rowcolor{top1-2} DR-mmst-APBF        & 0.965 & 0.959 & 0.954 & 0.965 & 0.964 \\ \hline
			\rowcolor{top1}   DR-mmst-APBF-e2e    & 0.972 & 0.98  & 0.964 & 0.959 & 0.98  \\
			\hline 
		\end{tabular}
	\end{table}
	
	\subsection{Baseline Comparison on Robustness}
	\label{subsec:rob_baseline}

\begin{figure*}[]
		\centering  
		\subfloat[JPEG]{
			\begin{minipage}[c][1\width]{0.20\textwidth} %
				\includegraphics[width=\textwidth]{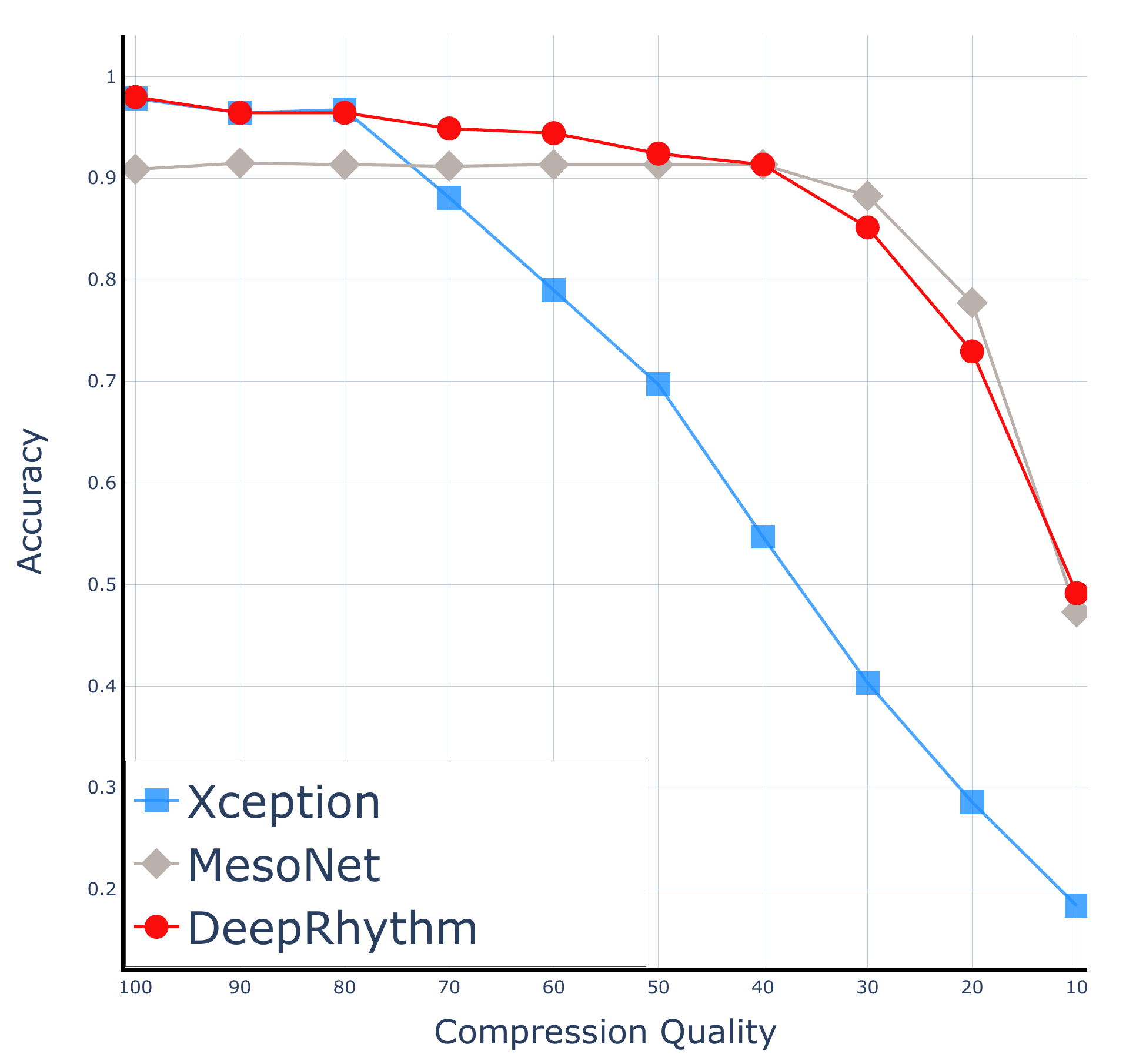}\\
				\includegraphics[width=1.0\textwidth]{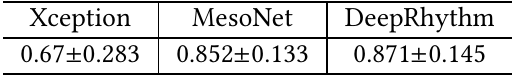}\\~
			\end{minipage}\label{fig_sub_jcom_base}
		}
		\subfloat[Blur]{
			\begin{minipage}[c][1\width]{0.20\textwidth} %
				\includegraphics[width=\textwidth]{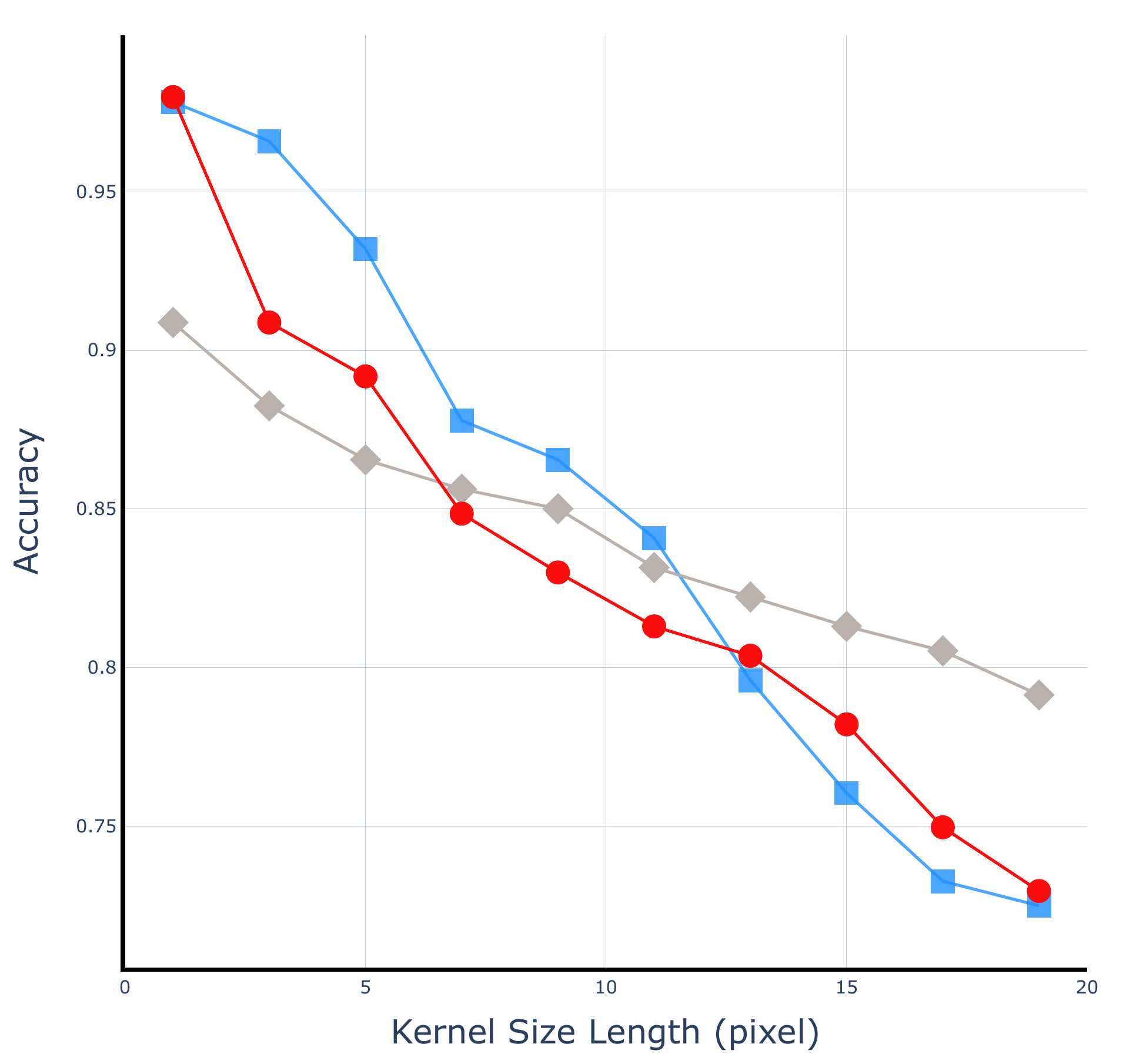}\\
				\includegraphics[width=1.0\textwidth]{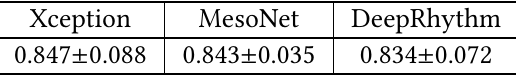}\\~
			\end{minipage}\label{fig_sub_blur_base}
		}
		\subfloat[Sampling]{
			\begin{minipage}[c][1\width]{0.20\textwidth} %
				\includegraphics[width=\textwidth]{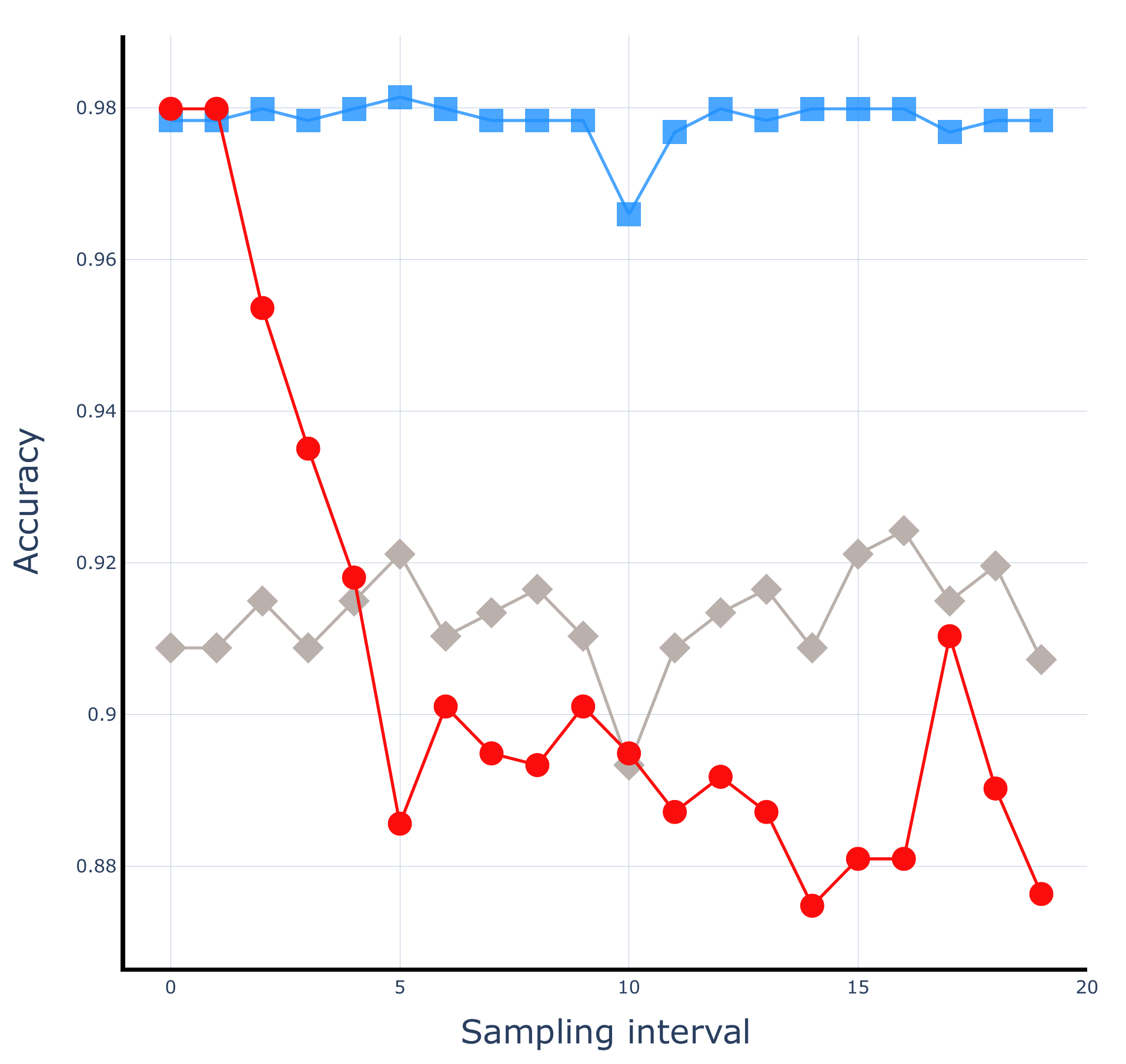}\\
				\includegraphics[width=1.0\textwidth]{./t2_blur.pdf}\\~
			\end{minipage}\label{fig_sub_frm_base}
		}
		\subfloat[Gaussian Noise]{
			\begin{minipage}[c][1\width]{0.20\textwidth} %
				\includegraphics[width=\textwidth]{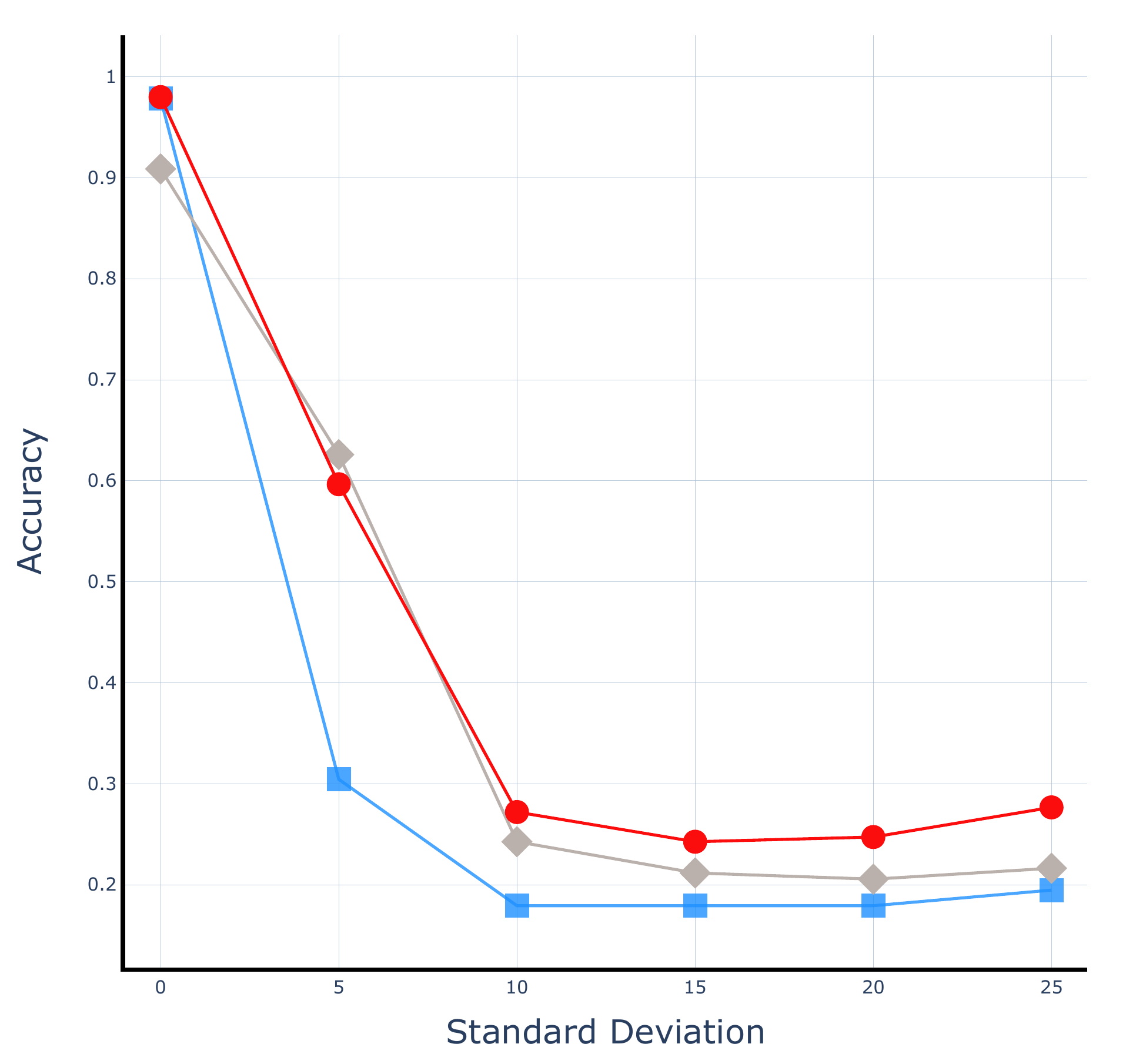}\\
				\includegraphics[width=1.0\textwidth]{./t2_blur.pdf}\\~
			\end{minipage}\label{fig_sub_noise_base}
		}
		\caption{Baseline comparison on robustness. We perform DeepFake detection through DeepRhythm and two state-of-the-art baselines, \ie, Xception and MesoNet, on a degradation dataset. Four degradations, \ie, JPEG compression, Gaussian blur, temporal sampling, and Gaussian noise, are added to the testing set of the ALL dataset. The average accuracy and corresponding standard deviation across all degradation degrees are presented at the bottom of each sub-figure.
		}
		\label{fig:rob_base}
	\end{figure*}

\begin{figure*}[]
		\centering  
		\subfloat[JPEG]{
			\label{fig_sub_jcom_abla}
			\includegraphics[width=0.20\textwidth]{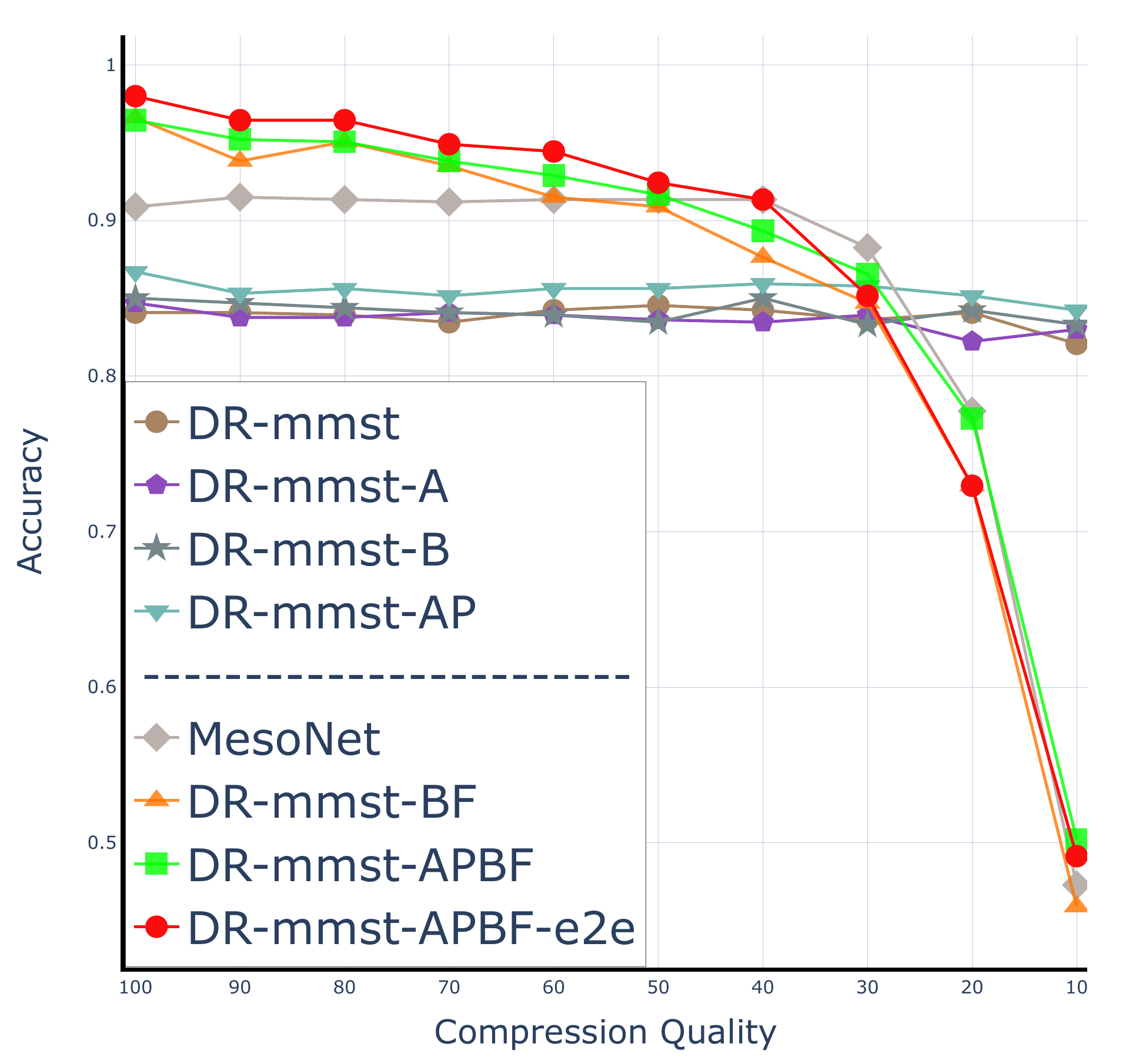}
		}\hspace*{-1em}
		\subfloat[Blur]{
			\label{fig_sub_blur_abla}
			\includegraphics[width=0.20\textwidth]{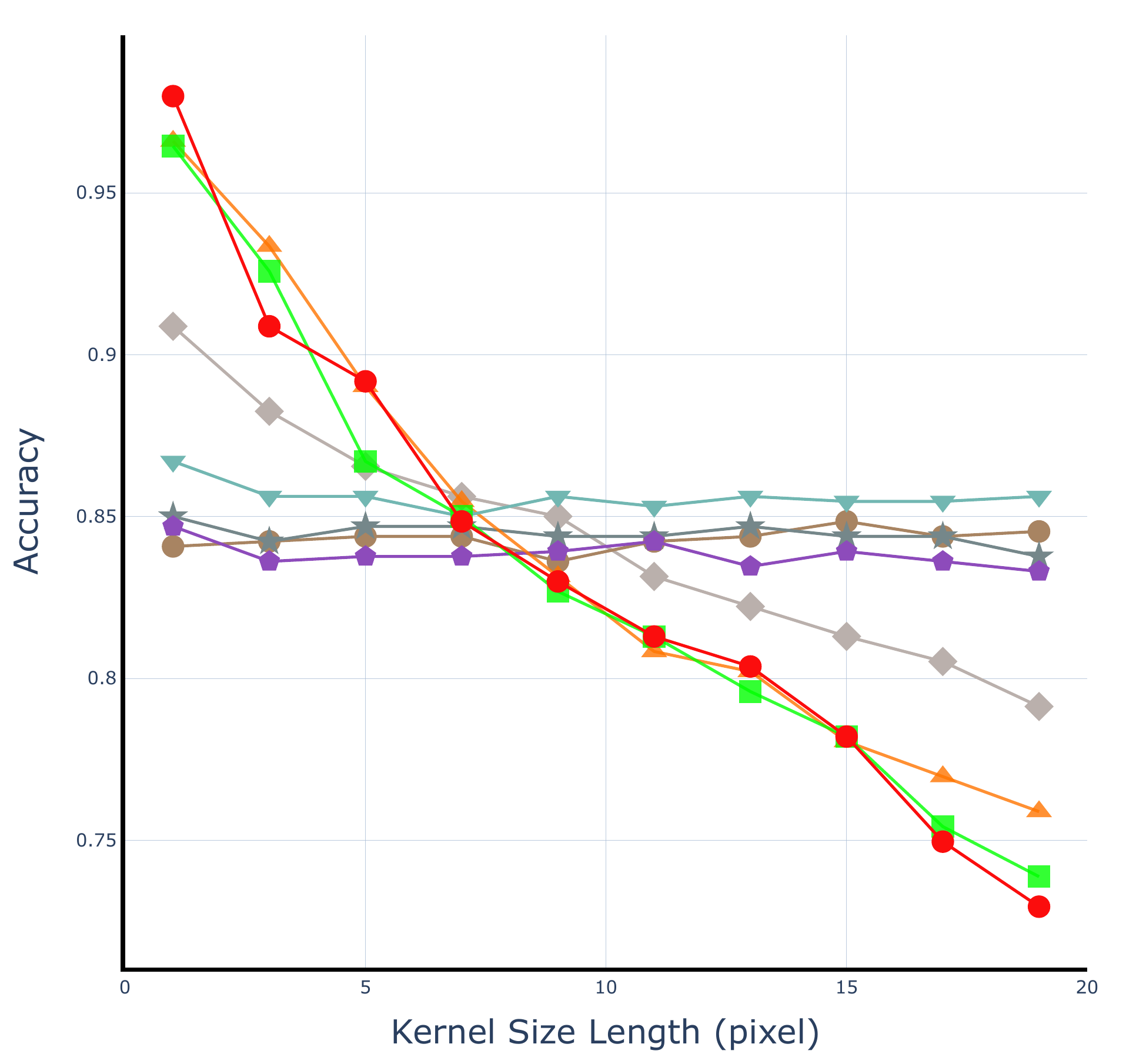}
		}\hspace*{-1em}
		\subfloat[Sampling]{
			\label{fig_sub_frm_abla}
			\includegraphics[width=0.20\textwidth]{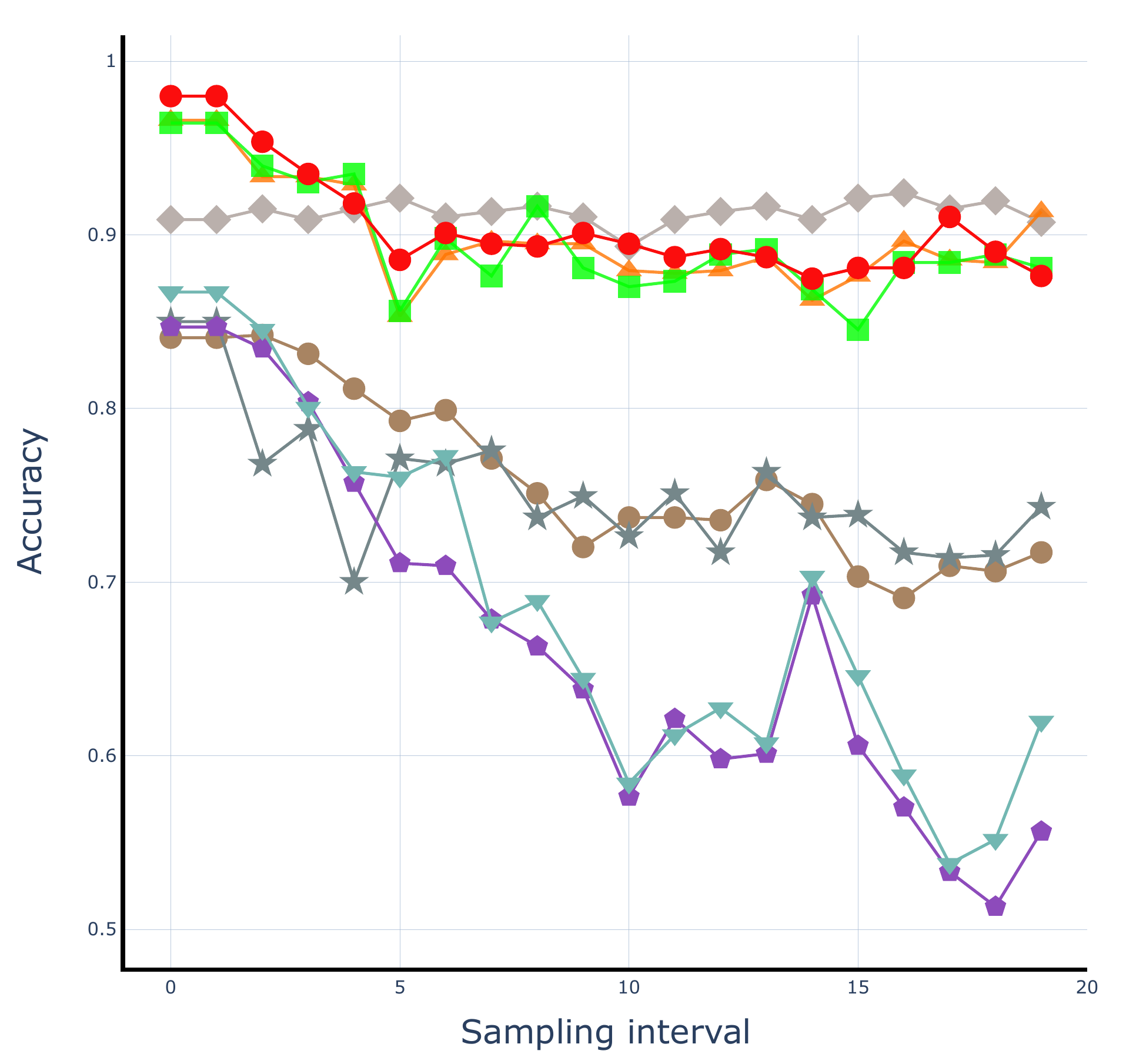}
		}\hspace*{-1em}
		\subfloat[Gaussian Noise]{
			\label{fig_sub_noise_abla}
			\includegraphics[width=0.20\textwidth]{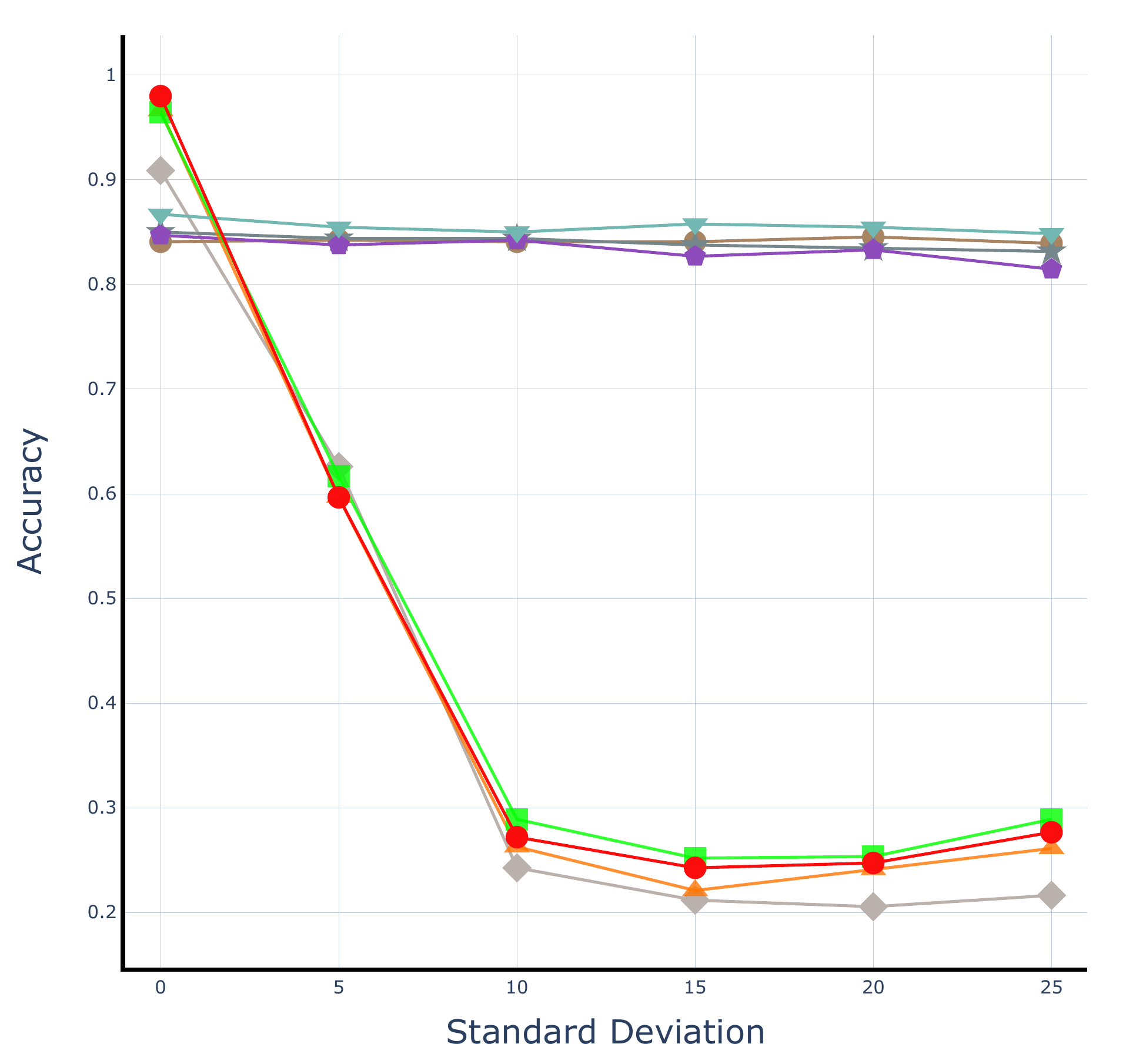}
		}\\ 
		\subfloat[Average accuracy~$\pm$~standard deviation across degradation degrees]{
			\label{rob_abla}
			\includegraphics[width=0.7\textwidth]{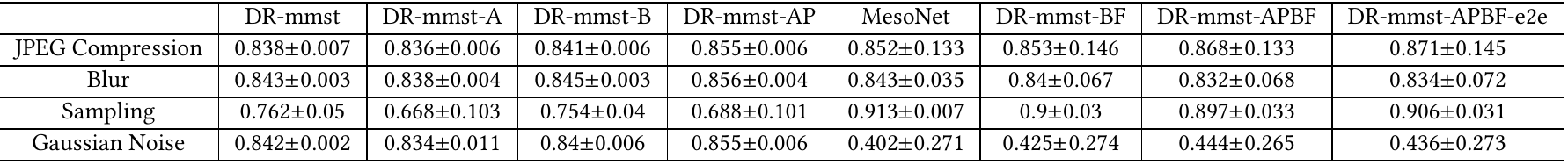}
		}
		\caption{Ablation Study on robustness. We perform DeepFake detection through MesoNet and DeepRhythm's seven variants on a degradation dataset. \if 0 Four degradations, \ie, JPEG compression, Gaussian blur, temporal sampling, and Gaussian noise, are added to the testing set of the ALL dataset. \fi The compared methods are clustered to two types, \ie, F-cluster using MesoNet for frame-level temporal attention (\ie, MesoNet itself, DR-mmst-BF, DR-mmst-APBF, and DR-mmst-APBF-e2e), and non-F-cluster that does not employ MesoNet (\ie, DR-mmst, DR-mmst-A, DR-mmst-B, and DR-mmst-AP). For each degradation, the average accuracy and corresponding standard deviation across all degradation degrees are presented at the bottom of figure.}
		\label{fig:rob_abla}
	\end{figure*}

	In this section, we study the robustness of our method and two baseline methods, \ie, Xception and MesoNet, which have the highest accuracy among baselines. Their models are trained on the training set of the ALL dataset. 
	We consider four general degradations, \ie, JPEG compression, Gaussian blur, Gaussian noise, and temporal sampling, and construct a degradation dataset by manipulating the testing set of the ALL dataset. For the first three degradations, we add the corresponding interference to each frame of the tested video and use the compression quality, blur kernel size, and standard deviation of noise to control the degradation degree, respectively. We show the degradation examples in Figure~\ref{fig:degrads}.
	The temporal sampling means that we do not use the raw continuous frames to get the MMST map but select frame at every $K$ frames. We use temporal sampling to test if our method still works under the unsmooth temporal variation. Please refer to the x-axis in Figure~\ref{fig:rob_base} and \ref{fig:rob_abla} for the variation range of each degradation. 


	As shown in Figure~\ref{fig:rob_base}, our method exhibits strong robustness on JPEG compression and Gaussian noise, but do not perform well on temporal sampling when compared with Xception and MesoNet. However, we could mitigate the issue with the video frame interpolation techniques, which is yet to be explored as the future work. 
	
	\subsection{Ablation Study on Robustness}

	We use the degradation dataset in Sec.~\ref{subsec:rob_baseline} to analyze the robustness of MesoNet (\ie, the frame-level temporal attention) and seven DeepRhythm variants (see the legend of Figure~\ref{fig:rob_abla}). These methods can be roughly divided into two clusters, one using MesoNet for the frame-level temporal attention (donated as F-cluster), including MesoNet, DR-mmst-BF, DR-mmst-APBF, and DR-mmst-APBF-e2e; the others do not employ the MesoNet (donated as non-F-cluster), including DR-mmst, DR-mmst-A, DR-mmst-B, and DR-mmst-AP.

	As shown in Figure~\ref{fig:rob_abla}, our MMSTR helps the variants, \ie, DR-mmst, DR-mmst-A, DR-mmst-B, and DR-mmst-AP, to keep at almost the same accuracy across all compression quality. The reason is that the MMSTR is calculated by average pooling pixel values in ROI blocks, thus is insensitive to local pixel variation caused by JPEG compression, Gaussian blur, and Gaussian noise. As shown in Figure~\ref{fig:degrads}, the MMST maps of JPEG compressed, noisy, and blurred videos are almost the same to the raw video. On the other hand, the MesoNet handles frames independently and relies on detailed information within frames. As a result, it helps our methods be robust to temporal sampling and achieve the best performance but is sensitive to local pixel variation.
	Clearly, the advantages and disadvantages of MMSTR and MesoNet are complementary. Our final version combining these two modules shows comprehensive robustness across all degradations.


	\section{Conclusions}\label{sec:conc}
	
	In this work, we have proposed DeepRhythm, a novel DeepFake detection technique. It is intuitively motivated by the fact that remote visual photoplethysmography (PPG) is made possible by monitoring the minuscule periodic changes of skin color due to blood pumping through the face. 
	Our extensive experiments on FaceForensics++ and DFDC-preview datasets confirm our conjecture that normal heartbeat rhythms in the real face videos are disrupted in a DeepFake video, and further demonstrate not only the effectiveness of DeepRhythm, but how it generalizes over different datasets by various DeepFake generation techniques.
	One interesting future direction is to study the combined effort of DeepRhythm with other DeepFake detectors \cite{ijcai20_fakespotter,arxiv20_deepsonar,arxiv20_fakepolisher,arxiv20_fakelocator}. Beyond DeepFake detection, the investigation of how DeepRhythm can be applied further to domains such as countering non-traditional adversarial attacks \cite{arxiv20_abba,arxiv19_amora,arxiv20_spark,arxiv20_pasadena} is also potentially viable. In addition, the possibility of using tracking methods \cite{Guo17_ICCV,Guo_TIP2020,DSARCF_TIP2019,Zhou17_ICASSP,SCT_TIP2017} to mine more discriminative spatial-temporal features would also be further studied.

	\begin{acks}
		This research was supported by 
		JSPS KAKENHI Grant No. 20H04168, 19K24348, 19H04086, JST-Mirai Program Grant No. JPMJMI18BB, Japan. 
		It was also supported by Singapore National Cybersecurity R\&D Program No. NRF2018NCR-NCR005-0001, National Satellite of Excellence in Trustworthy Software System No. NRF2018NCR-NSOE003-0001, NRF Investigatorship No. NRFI06-2020-0022, and the National Natural Science Foundation of China under contracts Nos. 61871258 and U1703261 and the National Key Research and Development Project under contracts No. 2016YFB0800403. We gratefully acknowledge the support of NVIDIA AI Tech Center (NVAITC) to our research.
	\end{acks}
	
	
	\balance
	
	\bibliographystyle{ACM-Reference-Format}
	\bibliography{ref}
	

\end{document}